%% file: main.tex
\pdfoutput=1
\documentclass{article}

    \usepackage[preprint]{neurips_2023}

\usepackage[utf8]{inputenc} % allow utf-8 input
\usepackage[T1]{fontenc}    % use 8-bit T1 fonts
\usepackage{hyperref}       % hyperlinks
\usepackage{url}            % simple URL typesetting
\usepackage{booktabs}       % professional-quality tables
\usepackage{amsfonts}       % blackboard math symbols
\usepackage{nicefrac}       % compact symbols for 1/2, etc.
\usepackage{microtype}      % microtypography
\usepackage{xcolor}         % colors

\usepackage{graphicx}
\usepackage{subfigure}
\usepackage{booktabs} % for professional tables
\usepackage{bbm}
\usepackage{enumitem,kantlipsum}

\usepackage{amsmath}
\usepackage{amssymb}
\usepackage{mathtools}
\usepackage{amsthm}
\usepackage{mdframed}
\usepackage{siunitx}
\usepackage{xcolor}
\theoremstyle{plain}
\newtheorem{theorem}{Theorem}[section]

\theoremstyle{definition}
\newtheorem{definition}[theorem]{Definition}
\newtheorem{assumption}[theorem]{Assumption}
\theoremstyle{remark}

\title{Black-Box Targeted Reward Poisoning Attack Against Online Deep Reinforcement Learning}

\author{%
  Yinglun Xu \\
  Department of Computer Science\\
  University of Illinois at Urbana Champaign\\
  \texttt{yinglun6@illinois.edu} \\
  \And
  Gagandeep Singh \\
  Department of Computer Science\\
  University of Illinois at Urbana Champaign\\
  \texttt{ggnds@illinois.edu} \\
}

\begin{document}

\maketitle

\input{000Abstract.tex}
\input{100Introduction.tex}

\input{200Related.tex}

\input{300Background.tex}

\input{400Methods.tex}

\input{500Experiments.tex}

\bibliographystyle{plainnat}
\bibliography{main}

\newpage
\appendix
\onecolumn
\input{600Appendix.tex}

\end{document}

%% file: 000Abstract.tex
\begin{abstract}
We propose the first black-box targeted attack against online deep reinforcement learning through reward poisoning during training time. Our attack is applicable to general environments with unknown dynamics learned by unknown algorithms and requires limited attack budgets and computational resources. We leverage a general framework and find conditions to ensure efficient attack under a general assumption of the learning algorithms. We show that our attack is optimal in our framework under the conditions. We experimentally verify that with limited budgets, our attack efficiently leads the learning agent to various target policies under a diverse set of popular DRL environments and state-of-the-art learners.
\end{abstract}

%% file: 100Introduction.tex
\section{Introduction}
Online deep reinforcement learning (DRL) algorithms have great potential to be applied in industrial applications such as robot control \citep{ChristianoLBMLA:17} and recommendation systems \citep{afsar2021reinforcement}. In such applications, the reward signals during training usually rely on human feedback. This raises the threat of training-time reward poisoning attacks. A user can deliberately provide malicious rewards to manipulate the DRL agent to learn a specific policy that has bad properties such as unsafe behavior to benefit the attacker. These policies can have high performance according to the agent's metric, making them harder to detect. Prior studies on attacks mainly focus on simpler tabular settings \citep{rakhsha2020policy,xu2021transferable,zhang2020adaptive, banihashem2022admissible}. The results show the vulnerability of current tabular RL algorithms, and the attacks only work in the white-box setting which may not be practical. In this work, we expose practical vulnerabilities of state-of-the-art DRL algorithms by finding a targeted black-box attack that works with limited budgets.

\textbf{Challenges for designing black-box targeted attack in DRL:} To make the attack realistic, we restrict the attacker to work in a black-box setting, that is, the learning algorithm and environment dynamics are unknown to the attacker. It only observes the interactions between the learning agent and the environment during training. We require the attacker to work with limited computational resources. Prior works in the simpler tabular MDP settings usually require knowledge of the environment and the learning agent \cite{xu2021transferable,zhang2020adaptive} which breaks the black-box constraint, and the complexity of the attack problem scale with the number of state and action spaces, which becomes impossible to solve in the continuous DRL setting. Furthermore, the computational resource required by the attacker can be more than that required by a learning agent, which may not be realistic. We need to find a black-box attack that requires limited computational resources while remaining efficient.   

\textbf{Our contributions:} 
To the best of our knowledge, our attack is the first black-box targeted attack for the DRL setting. We provide detailed theoretical analysis and experiments to show that our attack can successfully mislead the agent to a target policy with limited budgets and computational resources. To solve the challenges above, we first leverage a basic framework where the attack makes the agent training in a static adversarial environment by perturbing the true environment. By making general assumptions about efficient learning algorithms, we predict the general behavior of the agent during training without knowing the learning algorithm or predicting its exact behaviors. Next, we provide attack conditions to ensure efficient attacks, and then develop an optimal attack that can be constructed under the black-box setting and also satisfies the conditions with minimal budgets and very limited computational resources. Our attack is general and applicable to diverse environments and learning algorithms. We experimentally verify the efficiency of our attack on various popular DRL environments including MountainCar and HalfCheetah, and various state-of-the-art DRL algorithms including TD3 \citep{dankwa2019twin} and double dueling DQN \citep{wang2016dueling,van2016deep}. We consider various target policies to show that our attack works in a wide range of practical scenarios. Among all the attack settings we consider, our results show that our black-box attack can successfully make an agent take actions close to target actions during training with very limited budgets.

%% file: 200Related.tex
\section{Related Work}

\textbf{Data poisoning attack in simpler reinforcement learning}
Data poisoning attacks in reinforcement learning have been considered in simpler bandit cases \citep{jun2018adversarial,garcelon2020adversarial,liu2019data} and tabular MDP cases \citep{rakhsha2020policy,xu2021transferable,zhang2020adaptive, banihashem2022admissible}. Since the deep MDP case has a much more complicated environment to learn from, the attacks mentioned here cannot be applied to the deep MDP case. \cite{banihashem2022admissible, rakhsha2020policy} adopt the same basic attack framework as our work. Beside the difference in the definition of the attack's budget, the main limitation in their works is that their methods are only defined on discrete state and action spaces, and it also requires full knowledge of the environment. The attack problems proposed in the works are also impossible to solve in the deep MDP case due to the complexity of the environment.

\textbf{Observation perturbation attack in deep reinforcement learning}
There is a line of works study observation perturbation attack against DRL \citep{behzadan2017vulnerability, behzadan2017whatever, inkawhich2019snooping}. There are three main differences between these works and ours. First, observation perturbation attacks perturb the state signal, while our attack perturbs the reward signal. Second, observation perturbation attacks do not change the environment's dynamics but instead, change the  environment's signal observed by the agent. Our attack changes the dynamics of the environment. Third, the observation perturbation attack focuses on the intrinsic generalization vulnerabilities of the neural networks used by the learning agent, while our attack focuses on the vulnerabilities of the exploration strategy of the learning agent. 

\textbf{Data poisoning attack in deep reinforcement learning} \cite{xu2022efficient} study data poisoning for DRL but only for untargeted attacks. \cite{sun2020vulnerability} is the only other work that considers targeted attacks against DRL. There are three main limitations of this attack compared to ours and therefore less practical (a) the attack requires the knowledge of the learning algorithm (b) the attack only works for on-policy learning algorithms, and (c) the attacker decides perturbation after a batch of training steps. In addition, their attack problem is designed for minimizing the cost of corruption per training batch rather than each training step. Our attack is designed to minimize the per-step corruption and total corruption making our attack more general. In experiments, they only cover naive target policies which are not representative of practical attack scenarios. In our work, we experiment with multiple non-trivial types of target policies. We compare our work with \cite{sun2020vulnerability} in Appendix \ref{app:2}.

%% file: 300Background.tex
% \vspace{-0.05in}
\section{Background}
% \vspace{-0.05in}
First, we consider a standard online reinforcement learning setting consisting of an environment and a learning agent \citep{sutton2018reinforcement}. An environment is characterized by an MDP $\mathcal{M}=(\mathcal{S},\mathcal{A},\mathcal{P},\mathcal{R},\mu_0)$. Here $\mathcal{S}$ is the state space, $\mathcal{A}$ is the action space, $\mathcal{P}$ is the state transition function, $\mathcal{R}$ is the reward function, and $\mu_0$ is the initial state distribution. For deep reinforcement learning problems, the state space is always continuous. We study both discrete and continuous action spaces. Formally, the training process for the agent consists of multiple episodes. In each episode, the environment is initialized at a state $s$ sampled from $\mu_0$. The agent takes an action $a$, and the environment evolves to the next state $s'$ and returns an instant reward signal $r$ given by $\mathcal{P}(s,a)$ and $\mathcal{R}(s,a)$. When there is no poisoning attack, the agent will observe $(s,a,s',r)$, and continue this process until an episode is terminated. A policy $\pi$ for the environment is a mapping from a state $s$ to a probability distribution over the action space. For a deterministic policy, $\pi(s)$ is the action suggested by the policy. The policy value of a policy is the expected return in an episode when interacting with the environment $\mathcal{M}$ following the policy $\pi$. The goal of the agent is to find an optimal policy $\pi^*$ in the environment that has the highest policy value. We will use optimal action $\pi^*(s)$ to represent the action given by the optimal policy at state $s$.

Next, we consider an adversary which can inject perturbations to the reward during agent training. At each time step $t$, the current state is $s^t$, and the action taken by the agent is $a^t$. The true feedback of the environment is next state $s^{t+1}$ and reward $r^t$. Before the agent observes the true feedback $(s^t,a^{t+1},s^{t+1},r^t)$, the adversary observes it and injects perturbation $\Delta^t$ to the reward. Then, the agent receives $(s^t,a^{t+1},s^{t+1},r^t+\Delta^t)$. The general goal of the adversary is to make the agent learn a target policy $\pi^\dagger$ specified by the attacker. Note that an agent may only learn deterministic policies, so it is impossible for the attacker to make the agent learn a stochastic policy. Therefore, we only consider deterministic target policies and call the action $\pi^\dagger(s)$ given by the target policy as the target action at state $s$. We believe that this goal is natural and has been frequently applied in targeted attack communities. Next, we formally introduce the constraints, budgets, and goals of the attack.

\textbf{Constraints on the attack}:
We want the attack to work in a black-box setting with minimal information about the learner and environment. We also want the attack to work with limited computational resources. So we consider the following constraints on the attack:

\begin{mdframed}[backgroundcolor=gray!=10]
\begin{enumerate}[leftmargin=*]
    \item Oblivious to the learning agent: The attacker has no knowledge about the learning algorithms as well as the neural network used by the agent.
    \item Oblivious to the environment: The attacker has no knowledge about the transition function $\mathcal{P}$ and reward function $\mathcal{R}$ of the environment. It knows the state space $\mathcal{S}$ and action space $\mathcal{A}$.
    \item Limited computational resources: The attacker can observe the current state, action, and reward tuple $(s^t,a^t,r^t)$ at each timestep $t$ during training. However, it does not have significant memory or computing resources to store and learn based on historical observations.
\end{enumerate}
\end{mdframed}

\textbf{Budgets required by the attack}:
We want the attacker to work with limited budgets to ensure Stealthiness. In this work, we consider the following budgets for the attacker.

\begin{mdframed}[backgroundcolor=gray!=10]
\begin{enumerate}[leftmargin=*]
    \item The total amount of the reward perturbation $C$ during the training process the attacker applies. 
    \item The maximal amount of reward perturbation $B$ at each step. This ensures the perturbation at each step is bounded.
\end{enumerate}
\end{mdframed}

\textbf{The goal of the attack}:
In general, the attack wants to make the agent learn an arbitrary target policy. However, current learning algorithms do not specify a policy as the output when the training ends. In practice, people manually apply different criteria to specify the policy to output as the learned policy. Therefore, to objectively measure the efficiency of the learning algorithm, it is common to show the reward collected by the agent through the training process \citep{haarnoja2018soft}. Similarly, the efficiency of a targeted attack is evaluated through the similarity between actions taken by the agent and the target actions taken during training \citep{sun2020vulnerability,zhang2020adaptive}. We follow this convention and set the exact goal as follows:

\begin{mdframed}[backgroundcolor=gray!=10]
The goal of the attack is to induce a minimal difference between the actions selected by the agent and the target actions during training under the aforementioned constraints and budgets.
\end{mdframed}

We want to find an attack that works under the constraints and achieves the goal with limited budgets. In the next section, we will mathematically formalize the attacker problem statement, propose the attack method, and analyze its efficiency.

%% file: 400Methods.tex
% \vspace{-0.05in}
\section{Attack Framework and Methods}\label{sec:4}
% \vspace{-0.05in}

To mathematically formalize our attack goal, we define measures for the difference between actions.

\begin{definition}(Distance between two actions)
    For discrete action spaces, the distance between two actions $a_1$ and $a_2$ is defined as $d(a_1,a_2)=\mathbbm{1}\{a_1 \neq a_2\}$. For the continuous action spaces, the distance between $a_1$ and $a_2$ is defined as $d(a_1,a_2)=||a_1-a_2||_2/L$, where $L=\max_{x,y\in\mathcal{A}}||x-y||_2$ is the maximum $2$-norm distance between any two actions. 
\end{definition}

Note that in both discrete and continuous action spaces, $d(a_1,a_2)\in [0,1]$ always holds. Now we can formally define our attack problem. Let the total number of training steps be $T$. The perturbation at time $t$ be $\Delta^t$. The state and selected action at time $t$ during training under the perturbations be $s^t$ and $a^t$. Ideally, the optimal attack simultaneously minimizes: 1. the average distance $\epsilon$ between actions taken by the agent and the target actions 2. the total perturbation $C$, and 3. the maximum perturbation $B$ at each step. However, such an ideal optimal attack may not exist as an attack could achieve better $\epsilon$ for higher values of $B$ and $C$. Therefore, we define the optimal attack for a fixed budget of $B$ and $C$. The optimal attack here is the best attack one can find to make the agent take actions as close to the target actions as possible with given budgets. In this case, such an optimal attack must exist. The corresponding optimization problem for given values of $B$ and $C$ is defined as follows. 

% \vspace{-0.05in}
\begin{equation}\label{eq:optimal}
\begin{aligned}
    &\min_{\Delta^{t=1,\ldots,T}} \epsilon, s.t., \mathbb{E}[\sum_{t=1}^T d(a^t,\pi^\dagger(s^t))]/T = \epsilon,
    \mathbb{E}[\sum_{t=1}^T |\Delta^t|] = C, \max_t |\Delta^t| = B \\     
\end{aligned}
\end{equation}

 However, finding the optimal or even near-optimal attack is almost impossible under the black-box setting considered in our work. To find the optimal attack, the attacker needs to learn about the agent's algorithm and the environment $\mathcal{M}$ so that it can infer $a^t$ and $s^t$ to estimate the value of $d(a^t,\pi^\dagger(s^t))$. The only information for the attacker to determine the attack is the observation at each timestep, which is far from enough to approximate the dynamics of the environment. In addition, we do not want any specific assumption about the learning algorithms as we are looking for attacks that work for general learning algorithms. In this case, the relation between an algorithm's decisions at different time steps is unknown to the attacker, making it impossible to predict an algorithm's decisions in the future. So without the ability to approximate the environment and the learning algorithm, it is hard to find an optimal or near-optimal attack.

Due to the reasons above, we are only looking for attacks that are efficient. Formally, we define an attack to be $(\epsilon,C,B)$-efficient as follows:

\begin{definition}($(\epsilon,C,B)$ efficient attack)
    An attack is $(\epsilon,C,B)$ efficient if its perturbation $\Delta^{t=1,\ldots,T}$ is a solution to the following problem:
    % \vspace{-0.05in}
    \begin{equation} \label{eq:efficient}
    \begin{aligned}
        &\mathbb{E}[\sum_{t=1}^T d(a^t,\pi^\dagger(s^t))]/T = \epsilon,
        \mathbb{E}[\sum_{t=1}^T |\Delta^t|] = C, \max_t |\Delta^t| = B \\    
    \end{aligned}
    \end{equation}
    % \vspace{-0.1in}
\end{definition}

An attack is efficient if it satisfies Equation \ref{eq:efficient} with low values of $\epsilon$, $B$, and $C$. To infer the efficiency of an attack in this case, we adopt the following general assumption about a learning algorithm that is likely to hold in practice. Later we will design attack methods based on the assumption, and the result of the experiment shows the efficiency of the attack, which indirectly validates our assumptions.  

\begin{assumption}(efficient learning algorithm assumption)\label{asp:algorithm}
    Let the total training steps be $T$. We assume that for a given value of $\delta,p \in \mathbb{R}^+$, $\delta,p \ll 1$, there exists a set of environments $\mathbb{M}^\delta_p$ which we call `$(\delta,p)$-feasible set', such that for any efficient learning algorithm and any environment from the feasible set $\mathcal{M} \in \mathbb{M}^\delta_p$, after training for $T$ steps, the learning algorithm can guarantee that  with probability at least $1-p$, it will take the actions close to the optimal actions for most of the time during training, that is, $\sum_{t=1}^T \mathbb{E} [d(a^t,\pi^*(s^t))]/T \leq \delta$.
\end{assumption}

The assumption says that for many environments, an efficient learning algorithm should be capable to identify most optimal actions or actions close to them within $T$ steps with high probability. During training, it will take actions close to the optimal actions most of the time. We argue that the above assumptions are reasonable for efficient DRL algorithms. First, the goal of designing learning algorithms is to guarantee that the learning algorithm is able to find the optimal actions for a large set of environments. Second, a reinforcement learning algorithm needs to balance the exploration-exploitation trade-off, making it necessary to explore the sub-optimal actions in a strategic manner. In practice, most algorithms select the empirically optimal action with the highest probability such as $\epsilon$-greedy exploration \citep{van2016deep}. So the agent will identify the optimal actions or actions close to them as empirically best actions and correspondingly select those actions for most of the timesteps.

Next, we leverage a basic attack idea that utilizes the above assumption to construct efficient attacks. The idea is to make the agent train in a malicious environment $\widehat{\mathcal{M}}$ obtained by perturbing the true environment $\mathcal{M}$. In this case, since the agent is trained in a static environment under the attack, any assumptions about the algorithm learning in a static environment, including the above assumption, can hold. The corresponding attack framework we call `adversarial MDP attack' is defined below

\begin{definition}(Adversarial MDP attack)
    An adversarial MDP attack specifies an environment $\widehat{\mathcal{M}}$, which has the same state space, action space, and state transition function as the true environment, but a different reward function $\widehat{\mathcal{R}}$. The perturbation at time $t$ is $\Delta^t=\widehat{\mathcal{R}}(s^t,a^t)-\mathcal{R}(s^t,a^t)$. Under the attack, the agent will receive reward $\widehat{\mathcal{R}}(s^t,a^t)$ at time $t$. In another word, the agent is actually trained under $\widehat{\mathcal{M}}=(\mathcal{S},\mathcal{A},\mathcal{P},\widehat{\mathcal{R}})$ under the attack.
\end{definition}

Apparently, not all adversarial MDP attacks are efficient. We define efficient adversarial MDP sets for adversarial MDP attacks to set as $\widehat{\mathcal{R}}$ to ensure efficiency.  

\begin{definition}(Efficient adversarial MDP set)
    For an efficient adversarial MDP set $\text{EM}(\Delta,\delta,p)$ parameterized by $\Delta,\delta,p \in \mathbb{R}^+$, $\delta,p \ll 1$, any adversary MDP $\widehat{\mathcal{M}} \in \text{EM}(\Delta,\delta,p)$ satisfies:
    \begin{enumerate}[leftmargin=*]
    \item The adversary environment is in a $(\delta,p)$-feasible set $\widehat{\mathcal{M}} \in \mathbb{M}^{\delta}_p$ 
    \item The target policy $\pi^\dagger$ is the optimal policy under the adversarial environment $\widehat{\mathcal{M}}$.
    \item The differences between the adversarial and true reward functions are small for the actions whose distance to the target actions at any states. Formally, given a value of $\Delta \in \mathbb{R}^+$, for all state $s\in \mathcal{S}$, $|\widehat{\mathcal{R}}(s,a)-\mathcal{R}(s,a)| \leq \Delta \cdot d(a,\pi^\dagger(s))$.
    \end{enumerate} 
\end{definition}

The first condition says the adversarial environment is feasible for the agent to learn. The second condition says that the optimal actions in the adversarial environment are the target actions. The first two conditions ensure that the attack can utilize the efficient learning assumption and bound the average distance from selected actions to the target actions. The last condition requires similarities between the adversarial and true environments at the actions that are close to the target actions. This condition ensures that the perturbation required by the attack is small if the selected actions are close to the target action, which further bound the requirement on the budget of $C$. Later we will show a baseline attack whose adversarial environment follows the first two conditions but breaks the third. The attack requires a much higher value of $C$, resulting in a low performance. The theorem below shows that if an adversary MDP attack specifies an adversary MDP $\widehat{\mathcal{M}} \in \text{EM}(\Delta,\delta,p)$, then the attack's efficiency is guaranteed. The proof can be found in Appendix \ref{app:proof}. 

\begin{theorem}\label{thm:efficiency}
    Let the total training steps be $T$. If an adversarial MDP attack has $\widehat{\mathcal{M}} \in \text{EM}(\Delta,\delta, p)$, then it can satisfy equation $(1)$ with $\epsilon \leq \delta+p$, $B=\Delta$, and $C\leq (\delta + p) \cdot T \cdot \Delta$ 
\end{theorem}

% \vspace{-0.05in}
\subsection{Adaptive Target Attack}
% \vspace{-0.05in}

Theorem \ref{thm:efficiency} motivates us to focus on the efficient adversarial MDP set. It implies that an attack will be efficient if its $\widehat{\mathcal{M}}$ belongs to a $\text{EM}(\Delta,\delta,p)$  with minimal $\Delta$, so that it can be efficient with low values of both $B$ and $C$. This gives our main attack as below:

\begin{definition}(Adaptive target attack)
    The adaptive target attack is an instance of the adversarial MDP attack with $\widehat{\mathcal{M}}=\{\mathcal{S},\mathcal{A},\mathcal{P},\widehat{\mathcal{R}}\}$.  For a target policy $\pi^\dagger$, and a parameter  $\Delta \in \mathbb{R}^+$ , the reward function satisfies $\widehat{\mathcal{R}}(s,a)=\mathcal{R}(s,a)-\Delta \cdot d(a,\pi^\dagger(s))$.
\end{definition}

Note that the attack does not need any knowledge about $\mathcal{R}$ to apply corruption. It can select $\Delta^t=-\Delta \cdot d(a,\pi^\dagger(s))$ at round $t$. In the discrete action space, the attack penalizes the agent whenever it takes a non-target action and does nothing otherwise. In the continuous action space, the penalty $\Delta^t$ for an action $a^t$ is a linear function of its distance to the target action, that is $\Delta^t=\Delta \cdot ||a^t-\pi^\dagger(s^t)||_p/L$. The computational resource required by the attack is the memory to store the target policy $\pi^\dagger$, and the computational power to calculate the target action given by $\pi^\dagger$ and its distance to the action taken by the learning agent multiplied by a constant.

Next, we analyze why the adversarial environment of the adaptive target attack is a good choice among $\text{EM}(\Delta,\delta,p)$. Since no specific assumption is made about the learning algorithm, there is no information about the $(\delta,p)$-feasible environments, so we directly take the following assumption:

\begin{assumption}\label{ass:adaptive}
    For an adversarial MDP $\widehat{\mathcal{M}}$ constructed by the adaptive target attack, we assume there always exists $\delta,p \in \mathbb{R}^+$ such that $\widehat{\mathcal{M}}$ is in the $(\delta,p)$-feasible set of environments.
\end{assumption}
% \vspace{-0.05in}

In our experiments, we find that under the adaptive target attack with a small budget of $B$ and $C$, the learning agent gradually learns actions close to the target actions as the optimal ones, and the difference between the selected actions and the target actions decreases over time to a small value. This observation verifies our assumptions on the learning algorithm and the assumption that the adversarial environment constructed by the adaptive target attack is feasible. More details can be found in Appendix \ref{app:3}. To explain why this assumption holds in practice, we note that for discrete action spaces, the non-optimal actions are always associated with less instant reward, which intuitively makes it easier for the agent to learn the optimal actions. For continuous action spaces, the penalties incurred by the attack are continuous over the action space, so the perturbation signals are not sparse. Also, the perturbations have a coherent message that the closer to the target action an agent selects, the less instant penalty on reward is given. These features intuitively make the environment easier for the agent to learn the optimal actions.

Based on Assumption \ref{ass:adaptive}, the following theorem shows that the adaptive target attack is the adversarial MDP attack that can best highlight the target policy $\pi^\dagger$ to any learning agent.

\begin{theorem}
\label{thm:adaptive}
    The adversarial MDP of an adaptive target attack satisfies $\widehat{\mathcal{M}}= \arg \max_{\mathcal{M}' \in \text{EM}(\Delta,\delta, p)} \{\min_{\pi \neq \pi^\dagger}(V^{\pi^\dagger}_{\mathcal{M}'} - V^\pi_{\mathcal{M}'})\}$. 
\end{theorem}

% \vspace{-0.05in}
\subsection{Baseline attacks}
% \vspace{-0.05in}

Next, we provide some baselines for the adaptive target attack to compare against. Since no existing baseline can handle our black box setting, we design intuitive baseline attacks. These attacks are non-trivial with reasons to believe they may work, but we will point out their underlying problems and show in experiments that they are less efficient than the adaptive target attack.

\textbf{Greedy target attack.}
To highlight the necessity of the third adversarial attack condition that the adversarial and the true environments are similar in the actions close to the target actions, we provide a baseline attack called `greedy target attack'. For simplicity, we only define greedy target attack in the discrete action space. 

\begin{definition}(Greedy target attack)
    For a target policy $\pi^\dagger$, the reward function under the greedy target attack with a parameter $\Delta \in \mathbb{R}^+$ is $\widehat{\mathcal{R}}(s,a)=\mathcal{R}(s,a)+\Delta$ if $a = \pi^\dagger(s)$, and $\widehat{\mathcal{R}}(s,a)=\mathcal{R}(s,a)-\Delta$ if $a \neq \pi^\dagger(s)$.
\end{definition}

The greedy attack awards the agent for taking the target action and penalizes it otherwise. Although intuitively the attack makes the target actions more appealing, it has certain problems. First, it is not guaranteed that the target policy will be the optimal policy under $\widehat{\mathcal{M}}$ for a value of $\Delta$. Second, it needs to apply corruption at every round, resulting in a requirement on $C=\Delta \cdot T$, which is too large. To alleviate this issue, the greedy target attack can stop applying corruption early and hope that the agent will still believe the target policy to be optimal after. However, we will show through experiments that by using the same amount of $C$ with the same parameter $\Delta$, the adaptive target attack is always more efficient than the greedy target attack.

\textbf{Neighborhood target attack.}
To highlight the optimality of the linear reward penalty function given by the adaptive target attack in the continuous action space, we provide another baseline attack called `neighborhood target attack' for the continuous action spaces:

\begin{definition} (Neighborhood target attack)
    For a target policy $\pi^\dagger$ and an environment with a continuous action space, the adversarial reward function for the absolute neighborhood target attack with a parameter $r$ is defined as $\widehat{\mathcal{R}}(s,a)= \mathcal{R}(s,a) - \Delta\cdot\mathbbm{1}\{||a-\pi^\dagger (s)||_p > r\}$.
\end{definition}

The penalty function on reward under the neighborhood target attack is a step function. The attack can be thought of as an extension of the adaptive target attack in the discrete case by discretizing the continuous case. Although this seems simple and straightforward, there are some issues with this approach. First, even with a sufficient value of $\Delta$, one can only guarantee that the distances between the adversarial optimal actions and the target actions are no greater than $r$, so the choice of $r$ cannot be large. Second, it may be difficult for the agent to select an action within the neighborhood of the target action if $r$ is too small, so the choice of $r$ cannot be small which contradicts the previous concern. It could happen that no value of $r$ can address both concerns at the same time. Third, the absolute neighborhood target attack cannot directly utilize our efficient learning algorithm assumptions as the optimal actions are not target actions. In our experiments, we show that the performance of the neighborhood target attack, regardless of the choice on $r$, is much less than the adaptive target attack.

%% file: 500Experiments.tex
% \vspace{-0.05in}
\section{Experiment Results}
% \vspace{-0.05in}

We experiment on different environments learned by different learning algorithms to cover general cases. For the discrete case, we consider a popular classical control problem from Gym \citep{brockman2016openai}: MountainCar. For the continuous case, we consider a popular robot control problem from Mujoco \citep{todorov2012mujoco}: HalfCheetah. For the learning algorithms, we consider the state-of-the-art algorithms for discrete and continuous cases: double dueling DQN (for MountainCar), TD3 (for HalfCheetah). In Appendix \ref{app:2}, we will show the results for more environments and learning algorithms. For the target policies, to represent policies with diverse performances, we adopt the idea of the random, medium, and expert policies from \citep{fu2020d4rl}. A random policy is randomly generated, usually associated with low performance. An expert policy is a policy of the highest performance that an efficient learning algorithm can learn. A medium policy has a performance in the middle of the above two. The medium and expert target policies are generated by training in an environment with an efficient learning algorithm and saving the learned policies during training whose performances satisfy the aforementioned criteria. It is reasonable for the attack to consider high performing target policy, as the attack only cares about certain behavior of the target policy. Such target policies can still be malicious since their behavior may not be wanted by the learning agent. The number of total training steps for MountainCar is $\num{2.4e5}$, and the number for HalfCheetah is $\num{6e5}$. This choice ensures that when there is no attack, the algorithms can have high learning efficiency, i.e, they can collect high rewards in average during training. In addition to the baseline attacks proposed in Section \ref{sec:4}, in Appendix \ref{app:2} we show the result of the comparison to the va2c-p attack in \cite{sun2020vulnerability}.

We consider two scenarios on the budgets of the attack. In the first scenario, there is no hard limit on the attacker's budget, so the attack can always follow its strategy to apply perturbation, and we measure the attack's efficiency $(\epsilon,C,B)$ according to Equation \ref{eq:efficient} to verify our theoretical analysis in Section \ref{sec:4}. We will show that with a sufficient value of $\Delta$, the adaptive attack can have high efficiency with low values of $\epsilon$ and $C$ for any type of target policy. We will also show the efficiency of baseline attacks, but note that in this scenario it can be hard to say one attack is strictly more efficient than another. The reason is an attack may achieve a less value of $\epsilon$ but requires a higher value of $C$. Therefore, we consider the second scenario where we force a hard limit on the attacker's budget so that an attacker can never require a budget $C$ more than a specific value. This is also a realistic attack scenario, and all attacks will have the same values of $B$ and $C$, making it possible to compare the performance between attacks through the difference in $\epsilon$. We will show that our adaptive target attack has the highest performance in this case. For the results, each experiment is repeated for $10$ times, and we report the average result together with the confidence intervals.

\subsection{Efficiency of the adaptive target attack} 
% \vspace{-0.05in}
Here we measure the efficiency of the adaptive target attack with different fixed values of $\Delta$ according to our definition of attack's efficiency in Equation \ref{eq:efficient}. First, we want to show that given a sufficient value of $\Delta$, the adaptive target attack can have low values of $\epsilon$ and $C$. Second, we want to find the trade-off between $B$, $\epsilon$, and $C$ for the adaptive target attack. We directly have $B=\Delta$ by the definition of the attack, and we measure the empirical mean values of $\epsilon$ and $C$. The results are shown in Fig \ref{fig:delta}. The values of $\Delta$ are higher in the HalfCheetah environment because the unperturbed rewards there are higher. We find when the value of $\Delta$ is sufficient, the value of $\epsilon$ is small. When $\Delta$ becomes smaller, the value of $\epsilon$ becomes slightly larger. The reason can be the value of $\Delta$ still makes the target policy to be optimal, though the adversarial environment becomes harder to learn from. We also notice that sometimes when $\Delta$ is very small, the value of $\epsilon$ becomes close to the no-attack baseline, suggesting that in this case, the target policy is no longer optimal under the adversarial environment. These phenomena are also found theoretically and experimentally in tabular MDP cases \cite{zhang2020adaptive}. The results also show that the relation between $\Delta$ on $C$ is not monotonic. With a higher value of $\Delta$, the target actions are more likely to be learned by the agent as the optimal actions, therefore will be taken more frequently by the agent. However, a higher value of $\Delta$ also makes the perturbation at each timestep to be higher, so the value of $C$ may not decrease as the value of $\Delta$ increases.

\input{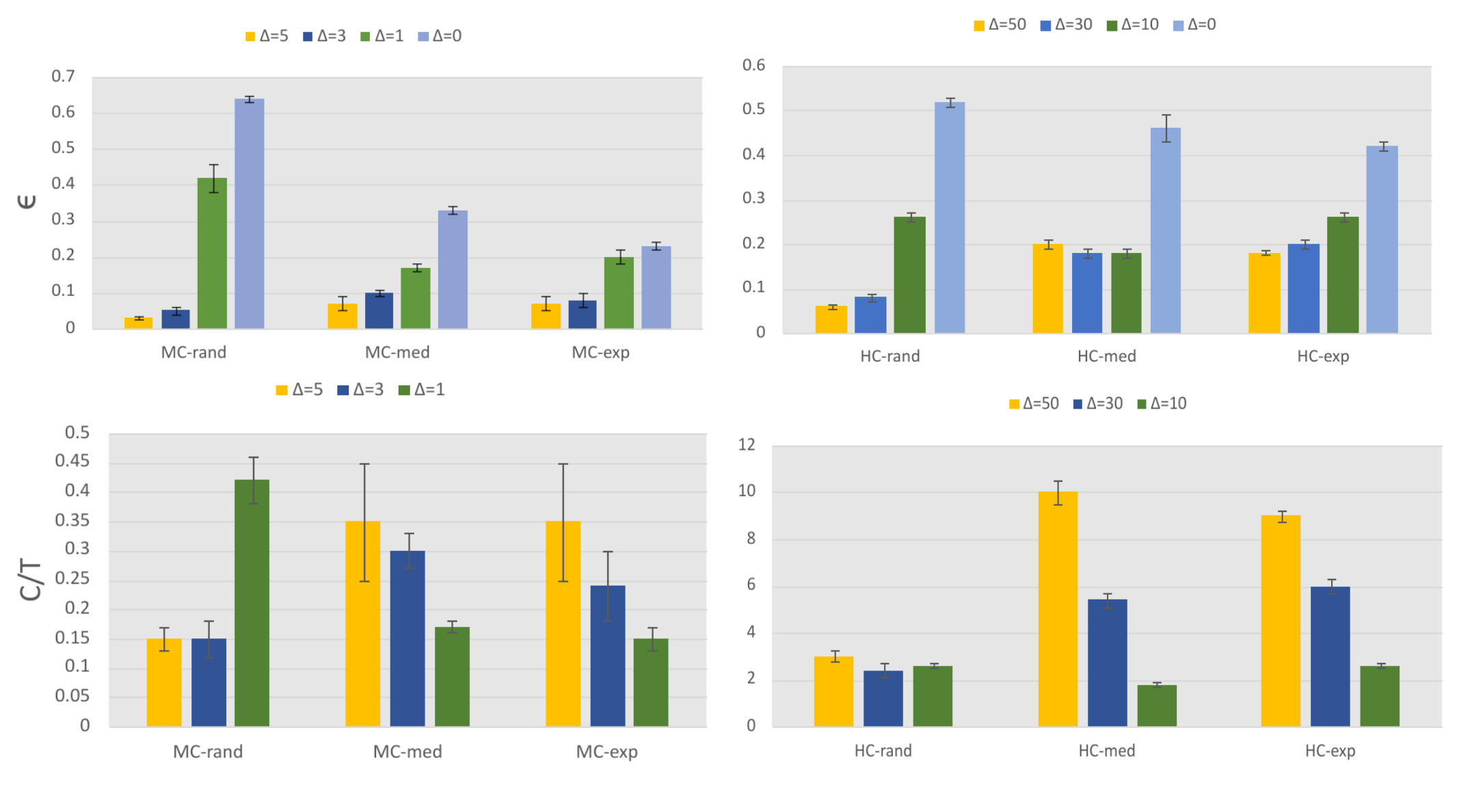}
\input{figures/baseline1}
\input{figures/baseline2}

In Fig \ref{fig:base1} we compare the adaptive target attack with the baseline attacks. For the baseline neighborhood target attack, as aforementioned, if $r$ is too large, the adversarial optimal policy may be too different from the target policy, so we test the attack with radius $r$ to be the same or twice the radius of the action space. We find that the neighborhood target attack is strictly less efficient than the adaptive target attack, as with the same value of $B$, it requires a higher value of $C$ and also induces a higher value of $\epsilon$. Compared to the adaptive target attack, the greedy target attack induces a similar value of $\epsilon$, but it requires a much higher value of $C$ which is unrealistic. To more clearly distinguish the difference in performance between attacks, we will compare the attacks in the second scenario where all attacks share the same values of $B$ and $C$. In that case, we show that the adaptive target attack has a much better performance than the greedy target attack.

In Appendix \ref{app:3}, we show the training process of the learning agent under the adaptive target attack with a sufficient budget of $C$ to further verify our understanding of the attack.

% \vspace{-0.05in}
\subsection{Performance of the adaptive target attack with fixed budget}
% \vspace{-0.05in}

Here, we evaluate the performance of the adaptive attack on misleading the agent to take actions close to target actions $\epsilon = \sum_{t=1}^T d(a^t, \pi^\dagger(s^t))/T$ with a hard limit on budget of $C$. In this case, the attack has to stop applying perturbation after the total perturbation reaches $C$. Recall in Section \ref{sec:4}, our analysis shows that with a sufficient value of $B$, the adaptive attack can achieve small values of $\epsilon$ and only requires a small budget of $C$. In addition, here we show that even when the budget is slightly less than the requirement of the attack, our adaptive attack can still make the agent mostly select target actions. The reason is that our attack already makes the agent learn most of the target actions as the optimal actions before running out of budget. So when there is no more perturbation, the agent will only seldom take actions far from the target actions, and it will take a long time for the agent to gather enough unperturbed new information to reverse the impact of previous corrupted information on the actions far from target actions. We test the performance of the attack with a fixed value of $B$ and different fixed values of $C$ by measuring the average distance between actions selected by the agent and target actions during training $\epsilon = \sum_{t=1}^T d(a^t, \pi^\dagger(s^t))/T$. The attack set $\Delta=B$ to utilize the constraint fully, and it stops applying perturbation when it runs out of budget $C$. We set $B=3$ for MountainCar and $B=50$ for HalfCheetah. The results are shown in Fig \ref{fig:base2}. We find that when the budget of $C$ is not too small, the adaptive attack is always able to achieve small values of $\epsilon$.

For the baseline greedy target attack, we find that its performance is less than our adaptive target attack. In addition to the fact that the greedy target attack is not efficient in terms of Equation 1, we have an intuitive explanation of why its performance is not well here. Since the greedy target attack will apply perturbation at every step, the attack will run out of budget $C$ at the beginning, and then the agent starts receiving unperturbed observations. Even if assuming the optimal policy in the adversarial environment to be the target policy, all the unperturbed observations contradict the previous corrupted observations, making it easier for the agent to correct its learned policy. We will also observe this phenomenon in the training logs shown later.
For the baseline neighborhood target attack, we test the attack with radius $r$ to be the same or twice as the radius of the action space and report the best result. We find that the attack has poor performance. This is not surprising as given a sufficient budget on $C$, the attack already induces a high value of $\epsilon$.

In Appendix \ref{app:3}, to intuitively show the behavior of the agent during training under the attacks in this scenario, we provide the training logs which record the average distance between selected actions and target actions at each training epoch.

\textbf{Conclusion and limitations:} 
In this work, we propose the first black-box targeted reward poisoning attack against online DRL. Our attack is practical because (1) it does not have knowledge of the learning agent and environment dynamics, (2) it requires limited budget for corrruption over each step and the whole training process, and (3) it requires limited computational resources. We hope our study will motivate the development of DRL algorithms robust against data poisoning attacks in the future. 
The limitations of our attack include: (1) only studies reward poisoning attacks (2) only works for online reinforcement learning settings.

%% file: figures/delta.tex
\begin{figure}[!ht]
\centering
\includegraphics[width=0.9\linewidth]{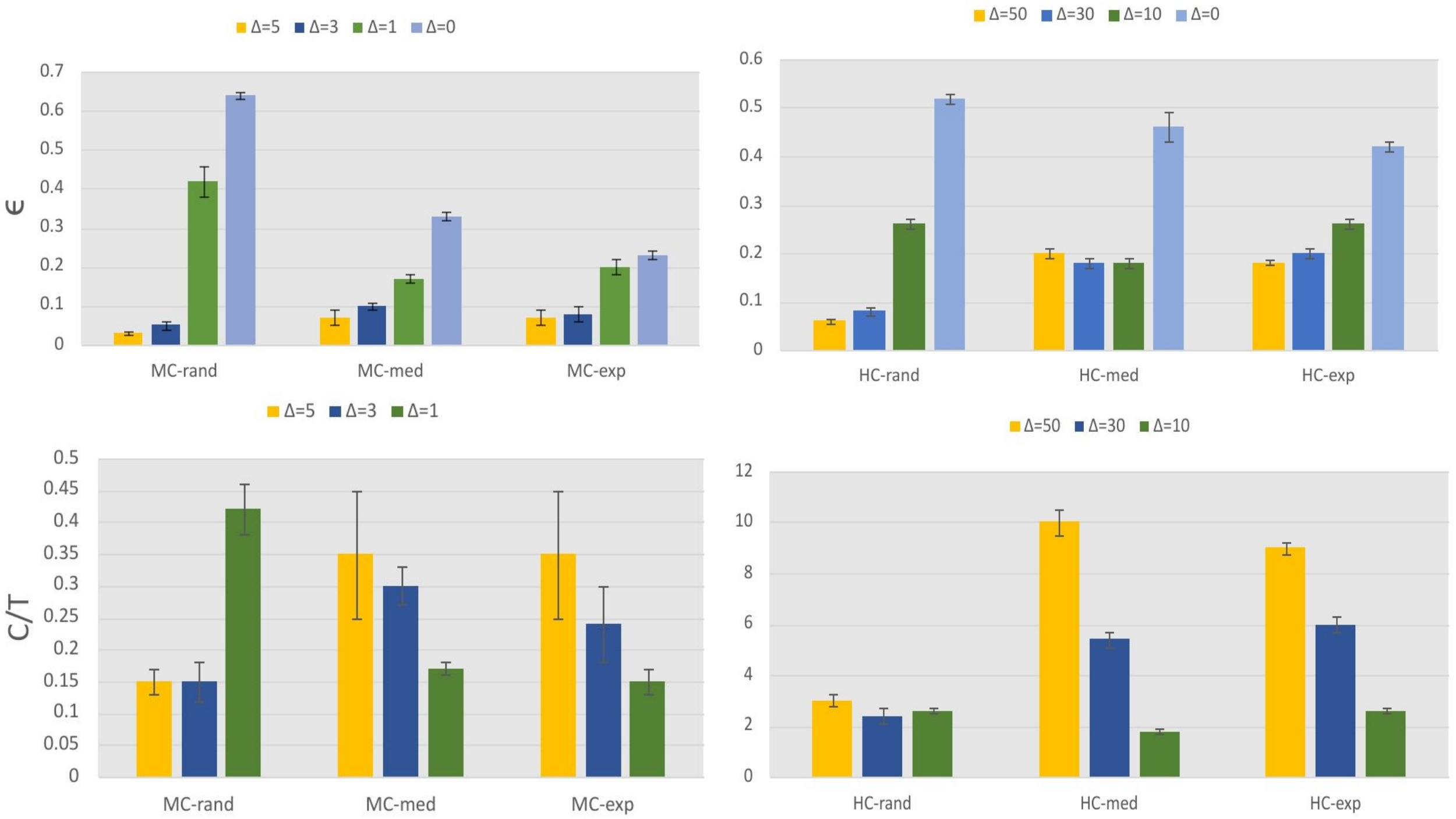}

\caption{Influence of $\Delta$ on $\epsilon$ and $C$ for the adaptive target attack. The two elements in the x-axis titles are the learning environment and target policy type. `MC' stands for `MountainCar', `HC' stands for `HalfCheetah', `rand' stands for `random', `med' stands for `medium', and `exp' stands for `expert'. $\Delta=0$ represents the clean case with no attack.}
\label{fig:delta}
\end{figure}

%% file: figures/baseline1.tex
\begin{figure}[!ht]
\centering
\includegraphics[width=0.9\linewidth]{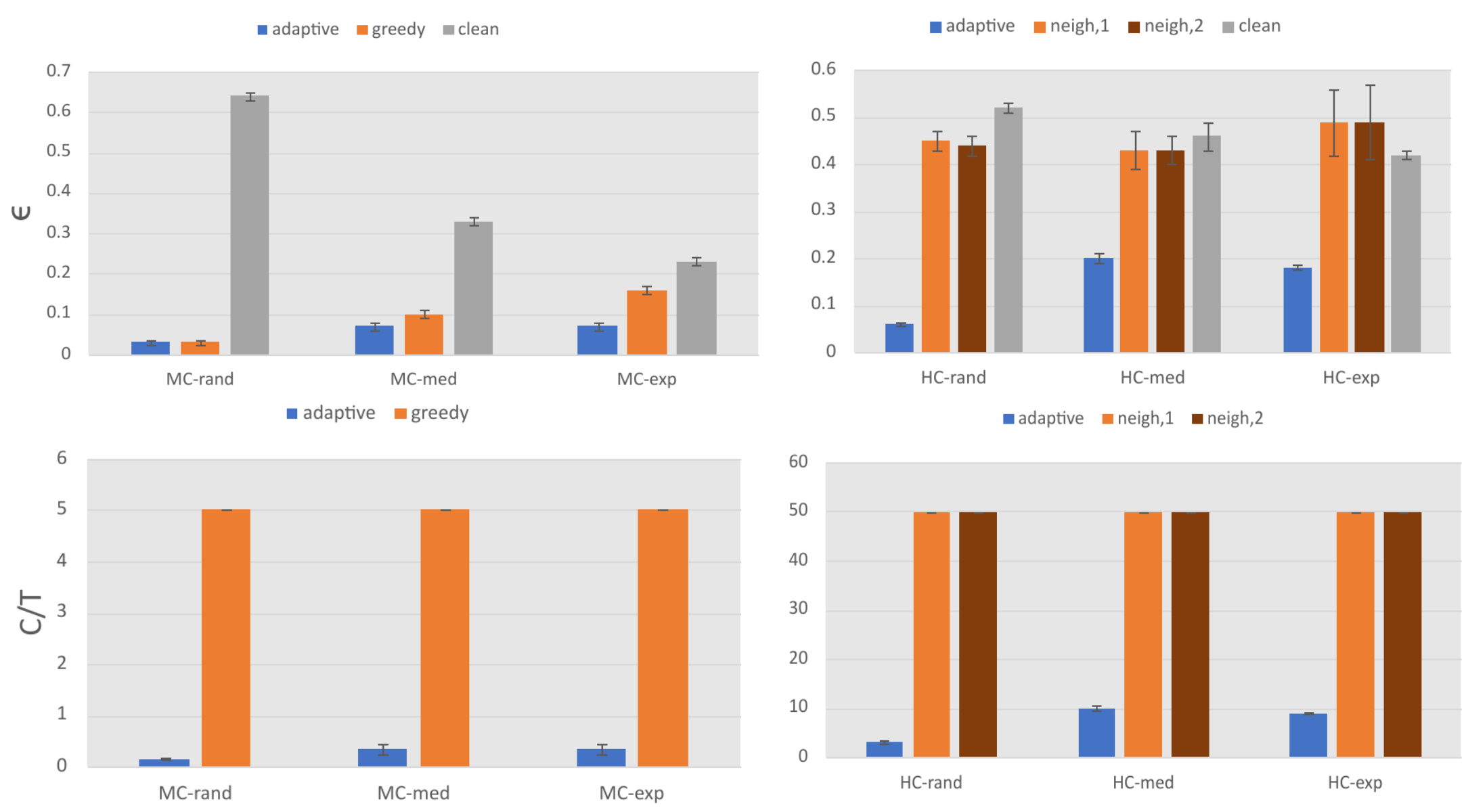}
\caption{Comparison between adaptive target attack and baseline attack with the same value of $|\Delta|$.  The two elements in the x-axis titles are the learning environment and target policy type.}
\label{fig:base1}
\end{figure}

%% file: figures/baseline2.tex
\begin{figure}[!htb]
\centering
\includegraphics[width=1.0\linewidth]{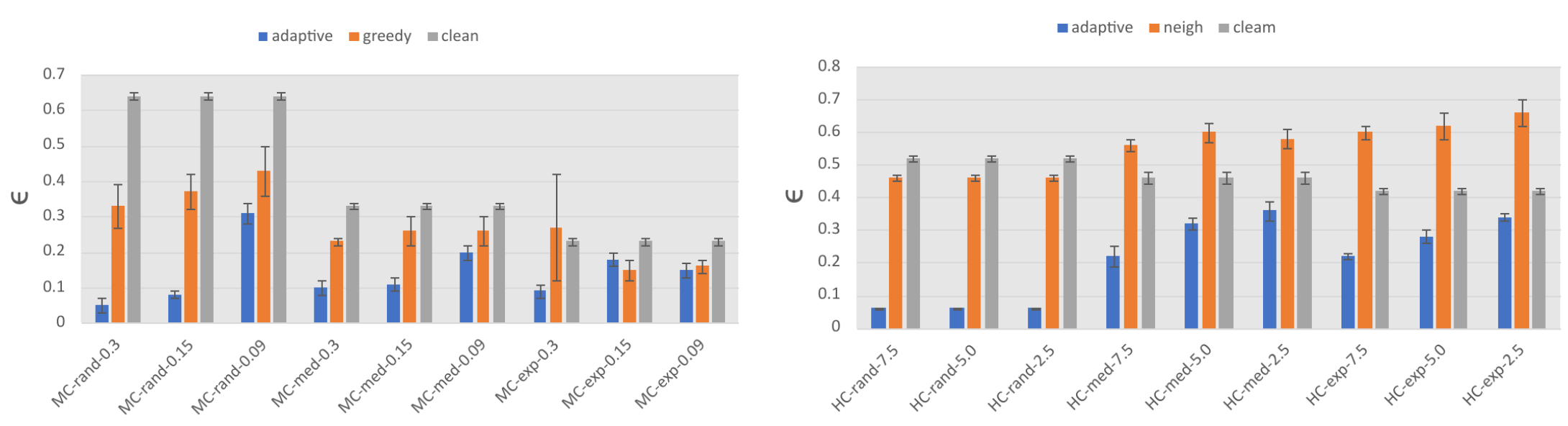}

\caption{Performance on $\epsilon$ of the adaptive target attack and baseline attacks with the same hard limit on $C$. The three elements in the x-axis titles are the learning environment, target policy type, and the value of $C/T$.}
\label{fig:base2}
\end{figure}

%% file: 600Appendix.tex
\section{Proof}\label{app:proof}
\textbf{Proof for Theorem \ref{thm:efficiency}}:
\begin{proof}
By the first efficient attack condition, the adversarial environment is $(\delta,p)$-feasible, so the guarantee on the agent's learning behavior given from Assumption \ref{asp:algorithm} can be applied. By the second efficient attack condition, we have $\hat{\pi}^*=\pi^\dagger$. Then together with the assumption $1$ from Assumption \ref{asp:algorithm}, the actions taken by the agent should satisfy $\sum_{t=1}^T d(a^t,\pi^\dagger(s^t))/T \leq \delta$ with probability at least $1-p$, otherwise $\sum_{t=1}^T d(a^t,\pi^\dagger(s^t)) \leq 1$ by definition. So we have $\epsilon \leq \delta+p$ for the first efficiency parameter of the attack.

Note that $d(\cdot,\cdot) \leq 1$ is always true, then by the definition of the attack, the perturbation at each step is $\Delta^t \leq \Delta \cdot d(a^t,\pi^\dagger(s^t) \leq \Delta$, suggesting that $B=\Delta$ for the second efficiency parameter of the attack.

By the third efficient attack condition and the definition of the attack, the total amount of perturbation is bound by $\sum_{t=1}^T \Delta^T \leq \Delta \cdot \sum_{t=1}^T d(a^t,\pi^\dagger(s^t) \leq \Delta \cdot (\delta + p) \cdot T$. So we have $C \leq \Delta \cdot \epsilon \cdot T$ for the third efficiency parameter of the attack. 
\end{proof}

\textbf{Proof for Theorem \ref{thm:adaptive}}:
\begin{proof}

Note that the policy value for a policy $\pi$ at a state $s_0$ can be decomposed as $V^\pi_{\mathcal{M}}(s)=E_{s\sim \mu^\pi(s_0)} \mathcal{R}(s,\pi(s))$, where $\mu^\pi(s_0)$ is the state distribution in an episode when following policy $\pi$ starting at state $s_0$. So the policy value for $\pi$ at state $s_0$ in the adversarial environment constructed by an attack that satisfies the third efficient attack condition can be written as 
\begin{equation}
\begin{split}
V^\pi_{\widehat{\mathcal{M}}}(s_0)&=E_{s\sim \mu^\pi(s_0)} \widehat{\mathcal{R}}(s,\pi(s))\\
&=E_{s\sim \mu^\pi(s_0)}\mathcal{R}(s,\pi(s)) \\
&- E_{s\sim \mu^\pi(s_0)} (\mathcal{R}(s,\pi(s))-\widehat{\mathcal{R}}(s,\pi(s)))\\
&\leq V^\pi_{\mathcal{M}}(s_0) - \Delta \cdot E_{s\sim \mu^\pi(s_0)} d(\pi(s),\pi^\dagger(s))\\
&:=V^\pi_{\mathcal{M}}(s_0) - \Delta \cdot D^{s_0}(\pi,\pi^\dagger)
\end{split}
\end{equation}

Here we denote $D^{s_0}(\pi_1,\pi_2)=E_{s\sim \mu^\pi_1(s_0)} d(\pi_1(s),\pi_2(s))$. Note that if $D^{s_0}(\pi_1,\pi_2)=0$, then $\pi_1$, $\pi_2$ generate identical trajectories starting at $s_0$ and can be considered as the same. The $\leq$ in the second last line can only take equality for the adaptive target attack given by the definition of the attack.

With the equation above, the difference between the policy value in $\widehat{\mathcal{M}}$ of the target policy and a policy $\pi$ at $s_0$ with $D^{s_0}(\pi,\pi^\dagger)>0$ is $$V^{\pi^\dagger}_{\widehat{\mathcal{M}}}(s_0)-V^\pi_{\widehat{\mathcal{M}}}(s_0) \leq (V^{\pi^\dagger}_{\mathcal{M}}-V^\pi_{\mathcal{M}})+\Delta \cdot D^{s_0}(\pi,\pi^\dagger).$$

Again the above equation takes equality only for the adaptive target attack. So the adversarial environment constructed by the adaptive target attack can maximize the gap between the performance of the target policy can any other policy under the adversarial environment.
\end{proof}

\section{Additional Experiments}\label{app:2}

In Table \ref{table:3}, we show the efficiency of the adaptive target attack in other environments learned by other algorithms. The additional environments we consider are Acrobot, CartPole, Hopper, and Walker2d. The additional learning algorithms we consider are dueling DQN \cite{wang2016dueling}, DDPG \cite{lillicrap2015continuous}, SAC \cite{haarnoja2018soft}.

\begin{table}[!ht]
\caption{Efficiency of the adaptive target attack in different environments learned by different learning algorithms}
\label{table:3}
\begin{center}
\begin{small}
\begin{tabular}{@{}lrrr@{}}
\toprule
Env-Tar-Alg-$\Delta$ & $\epsilon$ (clean) & $\epsilon$ (adaptive) & $C$ (adaptive)\\
\midrule
Acrobot-rand-double-$5$ & $0.68 \pm 0.01$ & $0.02 \pm 0.01$ & $0.10 \pm 0.05$\\
Acrobot-med-double-$5$ & $0.36 \pm 0.01$ & $0.08 \pm 0.01$ & $0.40 \pm 0.05$\\
Acrobot-exp-double-$5$ & $0.33 \pm 0.01$ & $0.09 \pm 0.01$ & $0.45 \pm 0.05$\\
CartPole-rand-duel-$5$ & $0.53\pm0.03$ & $0.11 \pm 0.08$ & $0.55 \pm 0.40$ \\
CartPole-med-duel-$5$ & $0.41\pm0.02$ & $0.23 \pm 0.06$ & $1.15 \pm 0.30$ \\
CartPole-exp-duel-$5$ & $0.44\pm0.02$ & $0.24 \pm 0.10$ & $1.20 \pm 0.50$ \\
\midrule
Hopper-rand-DDPG-$20$ & $0.42 \pm 0.04$ & $0.08 \pm 0.01$ & $1.6 \pm 0.2$\\
Hopper-med-DDPG-$20$ & $0.54 \pm 0.08$ & $0.17 \pm 0.01$ & $3.4 \pm 0.2$ \\
Hopper-exp-DDPG-$20$ & $0.52 \pm 0.02$ & $0.21 \pm 0.01$ & $4.2 \pm 0.2$ \\
Walker2d-rand-SAC-$30$ & $0.31\pm0.01$ & $0.10\pm 0.01$ & $3.0 \pm 0.3$ \\
Walker2d-med-SAC-$30$ & $0.50\pm0.02$ & $0.33\pm 0.02$ & $9.9 \pm 0.6$ \\
Walker2d-exp-SAC-$30$ & $0.54\pm0.02$ & $0.30 \pm 0.01$ & $9.0 \pm 0.3$ \\
\bottomrule
\end{tabular}
\end{small}
\end{center}
\end{table}

In Table \ref{table:4}, we show which algorithms are used to learn the target policies and the exact performance of the medium and expert target policies we choose. Note that the random target policies are randomly generated. 

\begin{table}[!ht]
\caption{Performance of the target policies used in the experiments}
\label{table:4}
\begin{center}
\begin{small}
\begin{tabular}{@{}lrrr@{}}
\toprule
Env-Alg & Medium & Expert\\
\midrule
MountainCar-duel & $-156$ & $-101$ \\
Acrobot-duel & $-199$ & $-100$\\
CartPole-double & $220$ & $500$\\
\midrule
HalfCheetah-DDPG & $5947$ & $12766$ \\
Hopper-TD3 & $1801$ & $3562$\\
Walker2d-TD3 & $2426$ & $4622$\\
\bottomrule
\end{tabular}
\end{small}
\end{center}
\end{table}

At last, we compare our attack to the va2c-p attack from \cite{sun2020vulnerability}. As an example, we consider the Swimmer environment learned by the PPO algorithm. To avoid degrading the effect of the va2c-p attack, we implement our attacks in their code and run the experiment. Due to the limitation in their code, only naive target policies are considered which always output the same action for any state. We adapt our attack to work with the constraints in their work by forcing the attack to stop applying perturbation if doing so breaks the constraint. For the values of constraints, we set $\epsilon=0.5$ and $C/K=1$. For our adaptive target attack, we set $\Delta=1$. The result of the training log of the agent under the attacks is shown in Figure \ref{fig:va2}. We find that our attack has comparable efficiency to the va2c-p attack. Note that the comparison is not fair in the first place since va2c-p attack works in white-box setting while our attack works in black-box setting. Our attack is also modified here as it is optimized for a different constraint on the attack that is actually more general. We also find that our attack runs about $50$ times faster than the va2c-p attack. It is noteworthy that our attack is also much easier to implement as its implementation is independent of the learning algorithm and only depends on the interactions during training.

\input{figures/comparison.tex}

\section{Training Logs under Different Attacks}\label{app:3}

In Fig \ref{fig:train_log}, we present the training log by showing at each training epoch, the average distance between the agent's actions and the target actions. The training log shows that the agent always gradually takes actions closer and closer to the target actions, suggesting that it gradually learns actions close to the target actions as the optimal ones. This observation agrees with our assumptions and analysis in Section \ref{sec:4}. In addition, we note that for medium and target policies, the average distance will also decrease as time goes on. The reason could be that there are certain similarities among policies of high performance.

\input{figures/training_log}

To intuitively show the behavior of the agent during training under the attacks in this scenario, we provide the training logs which record the average distance between selected actions and target actions at each training epoch. For each environment, we show the result for each type of target policy, and we only show the case where $C$ is the highest as it is the most powerful case considered in our experiments. The results are shown in Fig \ref{fig:base_log}. We find that under the adaptive target attack, the average distance always drops quickly as training goes on, and remains at a low value for a long time. The value may only increase slowly at a late time during training and remain below the result of the clean baseline. For the greedy target attack and neighborhood target attack, the general pattern for average distance versus training time is similar to that of the adaptive target attack, but the average distance always starts to increase much sooner, much faster, and sometimes even become larger than the value for the clean baseline, resulting in a big value of $\epsilon$ in the end. These observations directly suggest that the adaptive target attack is more efficient.

\input{figures/baseline_log}

%% file: figures/comparison.tex
\begin{figure}[!ht]
\centering
\includegraphics[width=0.5\linewidth]{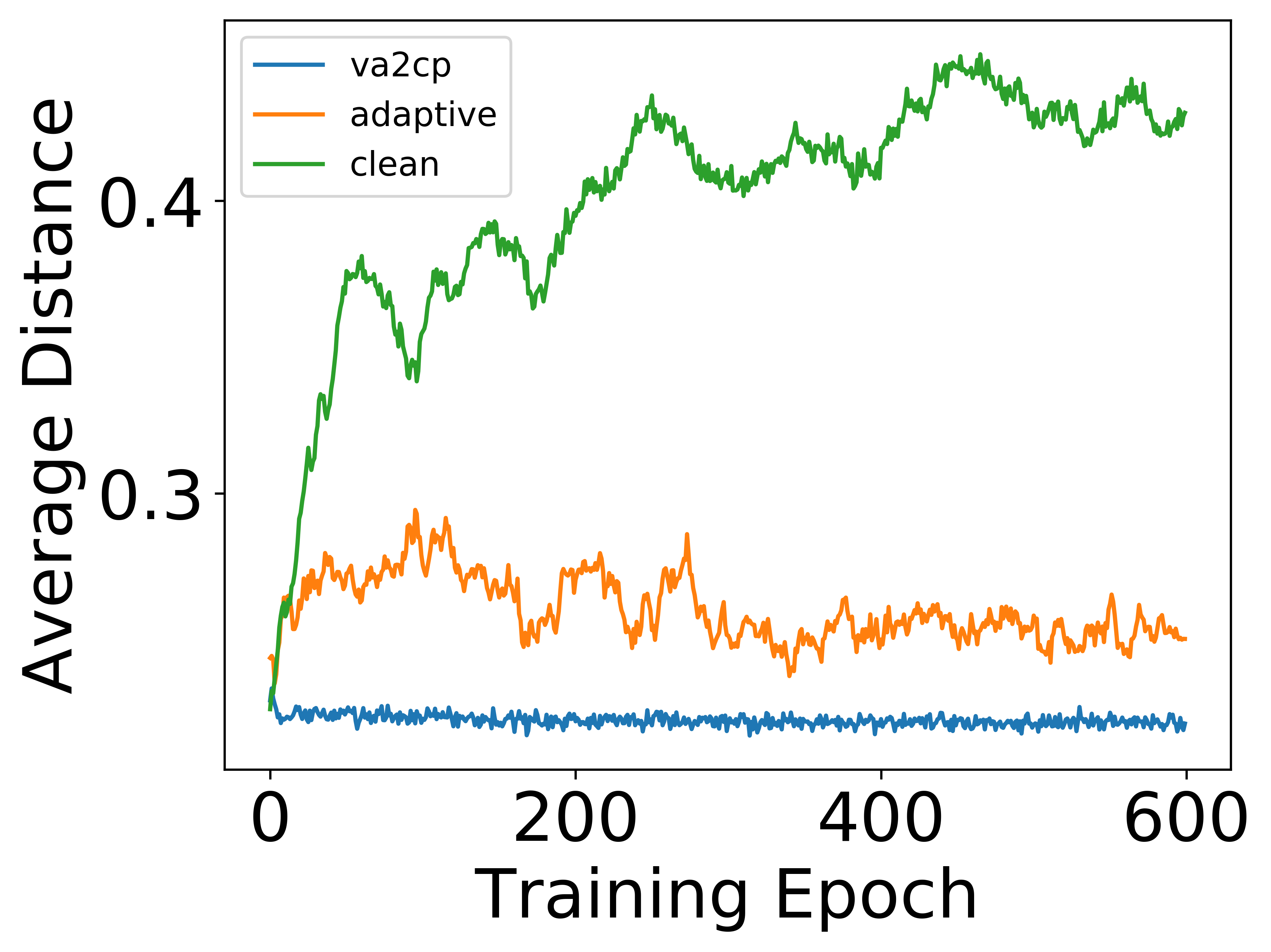}

\caption{Training log of PPO algorithm learned in Swimmer environment under the attacks.}
\label{fig:va2}
\end{figure}

%% file: figures/training_log.tex
\begin{figure}[!ht]
\centering
\includegraphics[width=1.0\linewidth]{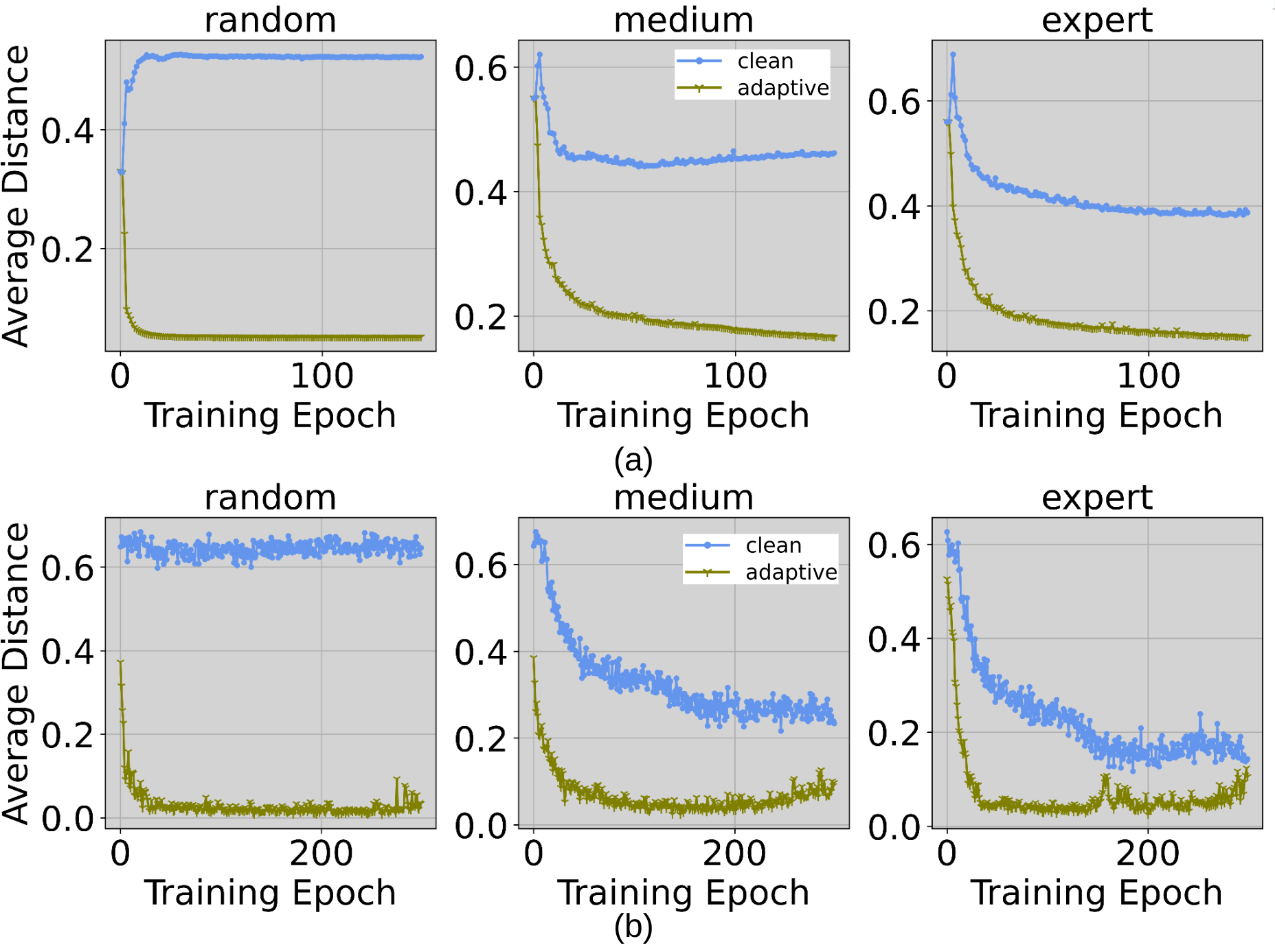}

\caption{Training log under no attack and adaptive target attack for (a) HalfCheetah environment with $\Delta=50$, (b) MountainCar environment with $\Delta=5$.}
\label{fig:train_log}
\end{figure}

%% file: figures/baseline_log.tex
\begin{figure}[!ht]
\centering
\includegraphics[width=1.0\linewidth]{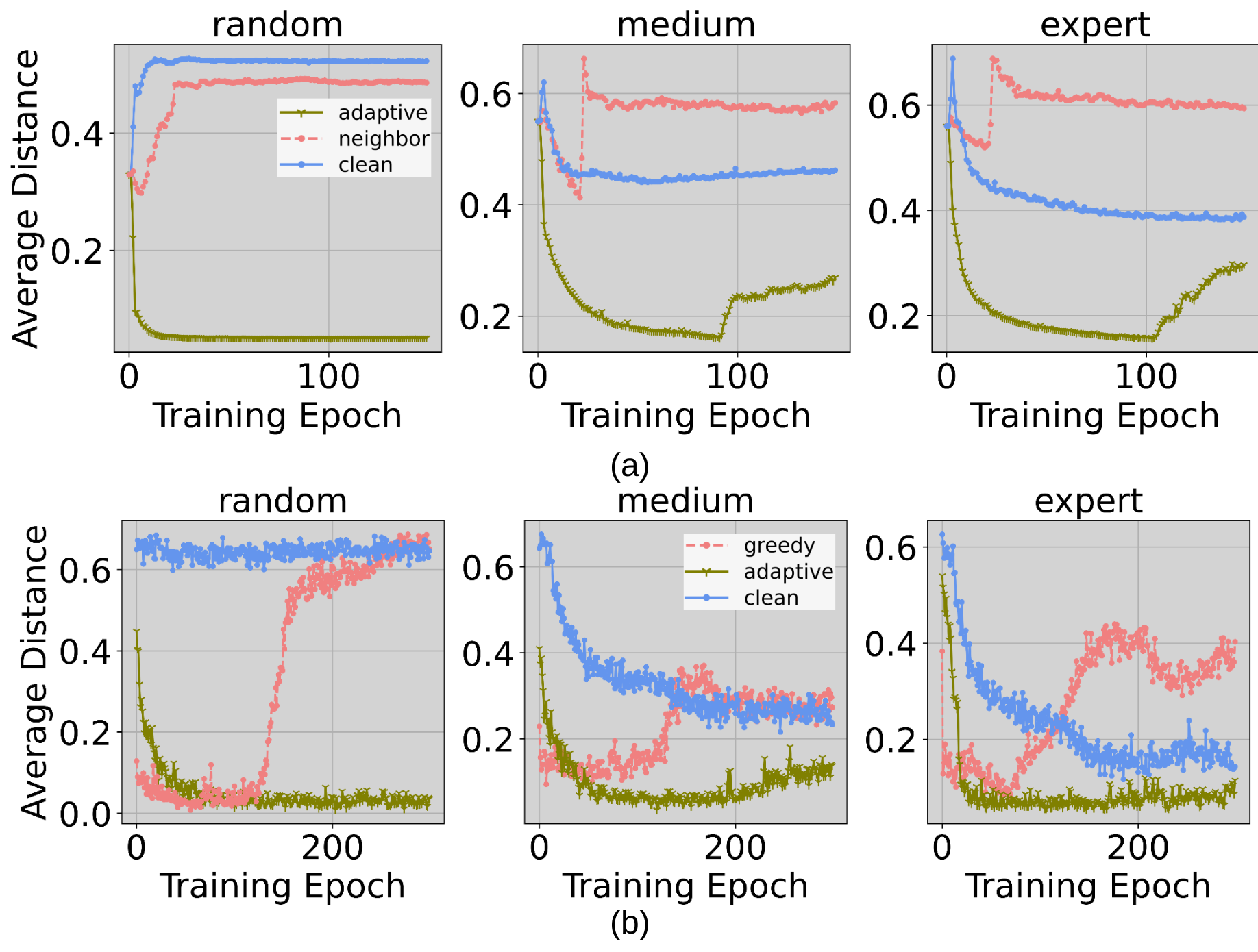}

\caption{Training log for adaptive target attack and neighborhood target attack in: (a) HalfCheetah  with a hard limit on budget $\frac{C}{T}=7.5$, and (b) MountainCar with a hard limit on budget $\frac{C}{T}=0.3$.}
\label{fig:base_log}
\end{figure}